\documentclass[letterpaper, 10 pt, conference]{ieeeconf}  
\IEEEoverridecommandlockouts                            

\usepackage{graphics} 
\usepackage{epsfig} 
\usepackage{graphicx}
\usepackage{xcolor} 
\usepackage{capt-of}

\usepackage{amsmath} 
\usepackage{amssymb}  
\usepackage{amsfonts}
\usepackage{mathtools}
\usepackage{units}

\usepackage{booktabs}
\usepackage{multirow}
\usepackage{tabularx}
\usepackage{multirow}

\usepackage{hyperref}
\usepackage{balance}
\usepackage{cite}

\usepackage{tikz}
\usepackage{pgfplots}
\usepackage{pgfplotstable}
\usepackage{circuitikz}
\usetikzlibrary{bending, plotmarks, shapes, pgfplots.statistics, pgfplots.colorbrewer, positioning}

\pgfplotsset{scaled y ticks=false}
\pgfplotsset{compat=1.16,
    width = 7cm,
    height = 3.5cm,
    title style = {yshift=-6pt},
    xlabel shift = -3pt,
    ylabel shift = -0pt,
    cycle list={{matlab1},{matlab2},{matlab3},{matlab9},{matlab4},{matlab5},{matlab6},{matlab7},{matlab8}},
    legend columns = 1,
    legend cell align={left},
    legend style ={
        draw = gray,
        fill opacity=0.8,
        text opacity=1.0,
        draw opacity=1.0,
    },
    xmajorgrids,
    ymajorgrids,
    scale only axis,
} 
\usepackage{xfrac}

\newcommand{\wfr}[0]{\ensuremath{\mathcal{W}}} 
\newcommand{\bfr}[0]{\ensuremath{\mathcal{B}}} 
\newcommand{\cfr}[0]{\ensuremath{\mathcal{C}}} 

\newcommand{\rotateRPY}[3]
{   \pgfmathsetmacro{\rollangle}{#1}
    \pgfmathsetmacro{\pitchangle}{#2}
    \pgfmathsetmacro{\yawangle}{#3}

    \pgfmathsetmacro{\newxx}{cos(\yawangle)*cos(\pitchangle)}
    \pgfmathsetmacro{\newxy}{sin(\yawangle)*cos(\pitchangle)}
    \pgfmathsetmacro{\newxz}{-sin(\pitchangle)}
    \path (\newxx,\newxy,\newxz);
    \pgfgetlastxy{\nxx}{\nxy};

    \pgfmathsetmacro{\newyx}{cos(\yawangle)*sin(\pitchangle)*sin(\rollangle)-sin(\yawangle)*cos(\rollangle)}
    \pgfmathsetmacro{\newyy}{sin(\yawangle)*sin(\pitchangle)*sin(\rollangle)+ cos(\yawangle)*cos(\rollangle)}
    \pgfmathsetmacro{\newyz}{cos(\pitchangle)*sin(\rollangle)}
    \path (\newyx,\newyy,\newyz);
    \pgfgetlastxy{\nyx}{\nyy};

    \pgfmathsetmacro{\newzx}{cos(\yawangle)*sin(\pitchangle)*cos(\rollangle)+ sin(\yawangle)*sin(\rollangle)}
    \pgfmathsetmacro{\newzy}{sin(\yawangle)*sin(\pitchangle)*cos(\rollangle)-cos(\yawangle)*sin(\rollangle)}
    \pgfmathsetmacro{\newzz}{cos(\pitchangle)*cos(\rollangle)}
    \path (\newzx,\newzy,\newzz);
    \pgfgetlastxy{\nzx}{\nzy};
}

\newcolumntype{C}{>{\centering\arraybackslash}X}
\newcolumntype{x}[1]{>{\centering\let\newline\\\arraybackslash\hspace{0pt}}p{#1}}
\definecolor{matlab1}{rgb}{0.00000,0.44700,0.74100}
\definecolor{matlab2}{rgb}{0.85000,0.32500,0.09800}
\definecolor{matlab3}{rgb}{0.92900,0.69400,0.12500}
\definecolor{matlab4}{rgb}{0.49400,0.18400,0.55600}
\definecolor{matlab5}{rgb}{0.4660, 0.6740, 0.1880}
\definecolor{matlab6}{rgb}{0.3010, 0.7450, 0.9330}
\definecolor{matlab7}{rgb}{0.6350, 0.0780, 0.1840}
\definecolor{matlab8}{rgb}{0.8, 0.8, 0}
\definecolor{matlab9}{rgb}{0.6, 0.6, 0.6}
\definecolor{verylightgray}{rgb}{0.98,0.98,0.98}

\pgfdeclarelayer{bg}   

\makeatletter
\g@addto@macro\@maketitle{
    \setcounter{figure}{0}
    \vspace*{12pt}
    \centering
    \includegraphics[width=17.6cm]{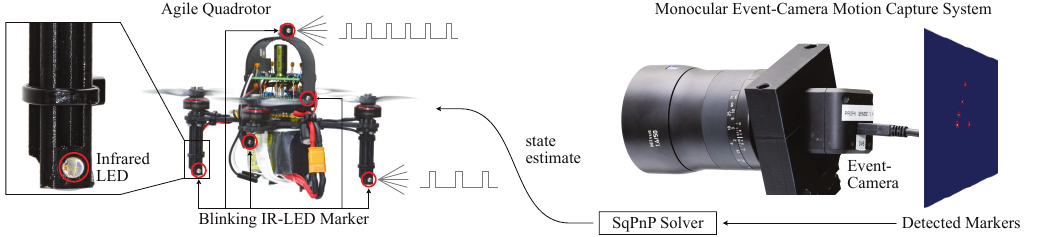}
    \vspace*{-3pt}
    \captionof{figure}{Overview of the monocular event-camera motion capture system: the object to track (e.g. a small quadrotor) is equipped with $N>=4$ infrared LED markers that blink at different frequencies. A single, calibrated event-camera is used to detect all markers and estimate the pose of the object by solving the perspective-n-points problem (PnP) with SqPnP~\cite{terzakis2020sqpnp}. The state estimate from this motion-capture system can then be used for closed-loop control.}
    \label{fig:overview}
    \vspace*{-12pt}
}

\title{\LARGE \bf
A Monocular Event-Camera Motion Capture System
}

\author{Leonard Bauersfeld and Davide Scaramuzza\\ 
Robotics and Perception Group, University of Zurich, Switzerland\\
\thanks{This work was supported by the European Union’s Horizon Europe Research and Innovation Programme under grant agreement No. 101120732 (AUTOASSESS), the European Research Council (ERC) under grant agreement No. 864042 (AGILEFLIGHT) and the Swiss National Science Foundation under project No. 200021\_212065.}
}

\begin{document}

\makeatletter
\maketitle
\thispagestyle{empty}

\begin{abstract}
    Motion capture systems are a widespread tool in research to record ground-truth poses of objects. Commercial systems use reflective markers attached to the object and then triangulate pose of the object from multiple camera views. Consequently, the object must be visible to multiple cameras which makes such multi-view motion capture systems unsuited for deployments in narrow, confined spaces (e.g. ballast tanks of ships). 
    In this technical report we describe a monocular event-camera motion capture system which overcomes this limitation and is ideally suited for narrow spaces. Instead of passive markers it relies on active, blinking LED markers such that each marker can be uniquely identified from the blinking frequency. The markers are placed at known locations on the tracking object. We then solve the PnP (perspective-n-points) problem to obtain the position and orientation of the object.
    The developed system has millimeter accuracy, millisecond latency and we demonstrate that its state estimate can be used to fly a small, agile quadrotor.
\end{abstract}
\section{Introduction}
\label{sec:introduction}
Motion capture systems are external camera systems that provide high-accuracy, low-latency pose estimates of an object by tracking markers attached to the object. The 3D position and orientation of the tracked object is then computed through multi-view triangulation. Such motion capture systems are ubiquitous in mobile robotics and they are used for platforms ranging from autonomous drones~\cite{ducard2009fma, kaufmann23champion}, remotely controlled (RC) cars~\cite{froehlich2022contextual}, and swimming vehicles~\cite{menolotto2020mocapreview} to manipulation~\cite{aljalbout2024transfer}. However, such motion capture systems require the object to be in view of multiple cameras, making deployment in narrow, confined spaces extremely difficult. Furthermore, the portability of the systems is negatively impacted by the long set-up time of the multi-camera system.

To overcome these drawbacks we developed the monocular, event-camera motion capture system shown in Fig.~\ref{fig:overview}. In contrast to multi-camera motion-capture systems that rely on triangulation, a monocular system must solve the PnP (perspective-n-point problem) problem~\cite{zisserman2004multipleview} to obtain the pose of the object in the camera frame. For a unique solution, at least four 3D~$\leftrightarrow$~2D correspondances must be known~\cite{gao2003complete}. Put differently, it is not enough to detect the markers, but the markers must also be uniquely identified. Note, in the PnP-setting it is always assumed that the locations of the markers on the tracking object are known, for example from a CAD model.

Event-cameras are an excellent tool to overcome the above limitations and build a monocular motion capture system. By using blinking LEDs as active markers the markers can be easily detected by the event-camera. To uniquely identify each marker, the LEDs blink at different, known frequencies~\cite{censi2013activeled}. The blinking frequency for each detected marker is then measured from the event stream to associate the detection in the image with an LED marker. The high temporal resolution of event cameras makes it possible to use very fast blinking frequencies and obtain a low-latency system, which is ideally suited for challenging real-world robotics tasks. 

The idea to use blinking LEDs in combination with an event camera for localization has been explored for over a decade~\cite{censi2013activeled, ebmer2024realtime6dof, loch2023eventbasedfiducial, salah2022neuromorphicvisionbased}, but most prior works focus on using the event camera on a mobile robot to localize w.r.t fixed markers. In this setting, the compute is constrained but markers can be large and consist of multiple LEDs. The availability of IMU measurements from the robot additionally simplifies the task when the event-camera is mounted the robot itself. Our system is different as it uses a static event camera and active markers on the robot, and as such presents a much improved version of~\cite{censi2013activeled} featuring true 6-DOF tracking, a 50-fold improvement in accuracy, and a 4-fold improvement in update rate. To the best of the authors knowledge, it is the first true monocular motion-capture system providing millimeter-accuracy 6-DOF pose estimates at update rates as high as \unit[1]{kHz}. This advance is enabled by modern sensors, an efficient, cache-friendly event-processing pipeline, utilizing a novel \emph{signed delta-time volume} event representation in combination with the very accurate and robust SqPnP~\cite{terzakis2020sqpnp} algorithm to estimate the pose of the tracked object.
\section{Related Work}
\label{sec:related_work}
The field of monocular pose estimation is mostly researched in the context of robot localization and approaches can be categorized into three families:
\begin{enumerate}
    \item standard cameras with fiducial markers (e.g. Aruco) 
    \item standard cameras with active point markers (e.g. LEDs) 
    \item event-cameras with active markers (e.g. blinking LEDs) 
\end{enumerate}

\subsection{Pose Estimation with Standard Cameras}
Most works on monocular pose estimation for mobile robots use standard cameras because those are readily available on most robots. With a standard camera, the common approach~\cite{breitenmoser2011monocular, benligiray2019stagstablefiducialmarker, kalaitzakis2021fiducial} is to use fiducial markers (e.g. AprilTag~\cite{olson2011apriltag}, Aruco~\cite{garrido2014aruco}) as such markers can robustly identified in RGB and grayscale images. 

Fiducial markers have a minimum size because the camera must still be able to detect the structure of the marker. Therefore, using infrared LEDs has been proposed as an alternative solution~\cite{faessler14monocular} to further miniaturize the system. The LEDs can be detected by thresholding an IR-filtered image, but in this setup the points can't be uniquely identified as all LEDs look the same to the camera. The pose estimation algorithm thus has to exhaustively search through all possible combinations.

Independent of the marker type, standard cameras are not well-suited for a low-latency monocular motion capture system as the latency of the overall system is strongly limited by the framerate of the camera. 

\subsection{Pose Estimation with Event Cameras}
In contrast to a frame-based camera an event-camera is inherently low-latency as events are output with microsecond latency. A major challenge is to design algorithms in a way that they are able to make use of this high update rate.

The first work~\cite{censi2013activeled} using event-cameras for localization\textemdash in a sense a direct predecessor of this work\textemdash uses blinking LEDs as active markers that can be identified based on their frequency. However, the pipeline is not able to estimate the location along the optical axis of the event-camera. Furthermore, it has noise levels on the order of \unit[10]{cm}, and thus is unsuitable for closed-loop control.

More recent monocular event-camera pose-estimation pipelines~\cite{salah2022neuromorphicvisionbased, ebmer2024realtime6dof} rely on very large active markers, up to $\unit[60\times 60]{cm}$ to achieve accurate localization. The markers are also containing multiple LEDs such that their identification is not just via frequency-detection, but similar to fiducial markers. 


\section{System Design}
\label{sec:system_design}

\begin{figure}
    \centering
    \input{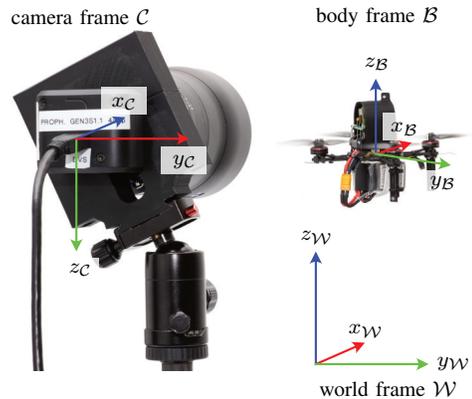}
    \caption{Definition of the camera frame ($z_\cfr$ is aligned with the optical axis), the body frame $\bfr$ and the world frame $\wfr$. The position of the active markers is defined in body-frame and must be known. The transform $T_{\cfr \bfr}$ is estimated through PnP and the pose of the camera in the world frame is assumed to be known.}
    \label{fig:coordinate_systems}
\end{figure}

This section gives a brief overview of the developed monocular event-camera motion-capture system. Details regarding the implementation of the blinking LED detection are given in Sec.~\ref{sec:implementation_details}. The blinking LED circuit itself is discussed in Sec.~\ref{sec:blinked_led_circuit}.

\subsection{Prerequisites}
In order to obtain accurate 3D pose estimation of the tracked objects, the event-camera must be calibrated and the location of the blinking LEDs must be known. Furthermore, the transform $T_{\mathcal{CW}}$ between the camera frame $\mathcal{C}$ and the world frame $\mathcal{W}$ must be known. The coordinate systems are illustrated in Fig.~\ref{fig:coordinate_systems}.

To calibrate the camera, we follow the approach from~\cite{muglikar2021calibrate} where the calibration is performed by first converting the event stream into event frames and then calibrating these using Kalibr~\cite{rehder2016extending}. The calibration also estimates the lens distortion and in this work we rely on a double-sphere distortion model because of its accuracy and computation efficiency~\cite{usenko2018doublesphere}. The event-camera is mounted horizontally on a stable tripod such that the transform from camera-frame $\mathcal{C}$ to world-frame $\mathcal{W}$ is fixed and known.

To ensure that the locations of the blinking LEDs are known and fixed, a 3D-printed LED holder is used. The locations of the LEDs are then directly known from the CAD model. The blinking frequency of each LED is also known, see Sec.~\ref{sec:blinked_led_circuit} for details on the circuit design.

\subsection{LED detection}
Robustly detecting blinking LEDs in an event-stream in real-time is the critical component of the motion-capture system. Events are processed in batches between \unit[1]{ms} (\unit[1]{kHz}) and \unit[2.5]{ms} (\unit[400]{Hz}), depending on the desired pose update rate. In a first step, all pixels whose event rate is below a threshold are discarded. The threshold is calculated based on the assumption that each LED period at least triggers two events and that a transition is detected with a given probability (e.g. \unit[80]{\%}).

For all pixels with a sufficient event rate, the average period and standard deviation is calculated. For details on this process, see Sec.~\ref{sec:implementation_details}. After identifying the average period, neighboring pixels with similar periods are clustered together. If the average period of the cluster is closer than $\unit[25]{\mu s}$ to one of the expected blinking frequencies from the LED, it is matched to that LED. The centroid of each LED is tracked with a constant-velocity particle filter.

\subsection{Pose Estimation}
To estimate the pose from the detected LED centroids, the centroids are undistorted first. Then, the PnP-problem is solved with SQPnP~\cite{terzakis2020sqpnp}. We chose this algorithm over other well-known algorithms such as EPnP~\cite{lepetit2009epnp} because SQPnP is fast enough and globally optimal~\cite{terzakis2020sqpnp}. 

The pose-estimate in the camera-frame is finally transformed into the world-frame and published via ROS. 
\section{Experimental Results}
\label{sec:experimental_results}
Often a motion capture systems are chosen because of their low noise levels and low latency which is ideally suited for robot control. Therefore, we demonstrate the performance of the developed system, by flying the small drone depicted in Fig.~\ref{fig:overview} in closed-loop with the event-camera motion capture being the only source of state estimation. We use a Prophesee Gen 3.1 event-camera ($\unit[640\times 480]{px}$, \unit[3/4]{inch} sensor) with either a \unit[25]{mm} or a \unit[50]{mm} lens resulting in a horizontal FoV (field of view) of \unit[22]{deg} and \unit[11]{deg}, respectively.

\subsection{Pose Estimation Noise}
\begin{figure}
    \centering
    \input{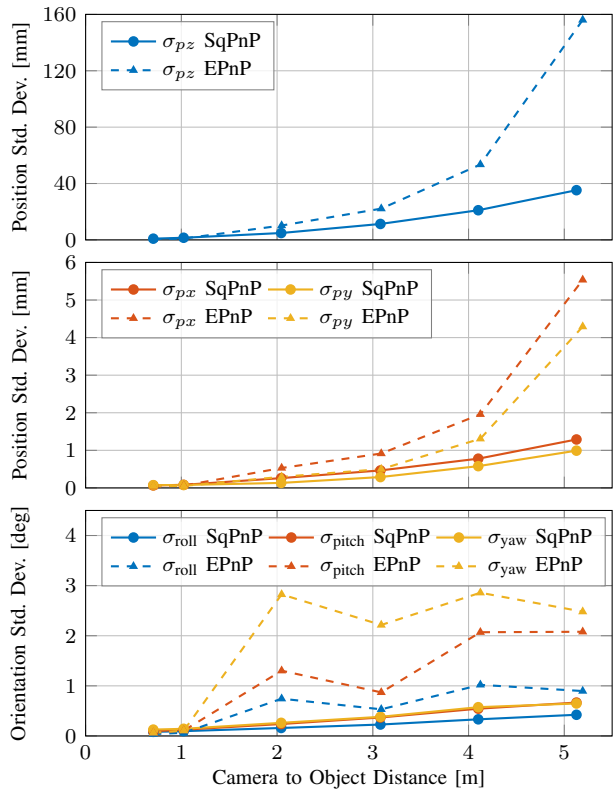}
    \caption{The object is placed at distances between \unit[70]{cm} and \unit[5]{m} statically in front of the camera (with the \unit[25]{mm} lens). The plots show the standard deviation in the position measurement ($z_\cfr$ and $x_\cfr$, $y_\cfr$) as well as the orientation measurements. We can clearly see that SqPnP~\cite{terzakis2020sqpnp} outperforms EPnP~\cite{lepetit2009epnp} by a large margin.}
    \label{fig:pose_noise}
\end{figure}

In a first experiment the drone is rigidly placed at various distances in front of the camera. Then, 10 seconds of data are recorded with the event-camera and we calculate the standard deviation of the pose estimate. Since the drone is static, all deviations from the mean are only due to noise. The results of this analysis summarized in Fig.~\ref{fig:pose_noise}. A few interesting observations can be made which are subsequently discussed in detail:
\begin{enumerate}
    \item The noise levels along the $z_\cfr$ axis are much larger compared to the $x_\cfr$ and $y_\cfr$ axis.
    \item The SqPnP~\cite{terzakis2020sqpnp} algorithm achieves a much better performance than EPnP~\cite{lepetit2009epnp}.
    \item For SqPnP the noise levels in position scale quadratically with the distance from the camera, while the orientation noise scales linearly.
\end{enumerate}

The $z_\cfr$-axis is the optical axis of the camera and hence the $z_\cfr$ coordinate can only be inferred from the scale of the object. Assuming that elongation of the object along the optical axis is small compared to the distance to the camera (i.e. the object is nearly flat), a well-known result from stereo-vision applies~\cite{zisserman2004multipleview}: for a given inter-marker distance $d$, focal length $f$ and a marker detection with uncertainty $\sigma_u$ (in pixels) the depth uncertainty $\sigma_{pz}$ scales with the square of the distance $z$ as
\begin{equation}
    \sigma_{pz} = \frac{\partial z_\cfr}{\partial u} \cdot \sigma_u = \frac{b \cdot f}{z_\cfr^2} \cdot \sigma_u \sim \frac{1}{z^2}\;.
\end{equation}
For the positional errors in $x_\cfr$ and $y_\cfr$ we observe a similar quadratic dependency on the camera-object distance, however with much less noise. Intuitively, this makes sense as the translation along $x_\cfr$ and $y_\cfr$ are directly observable from each marker and thus the estimate is much more accurate.

The position noise plots also highlight the superior performance of SqPnP for this task: the optimization-based approach is able to estimate the position with much less variance given the same input data. This discrepancy becomes even larger when considering the orientation estimation shown at the bottom plot of Fig.~\ref{fig:pose_noise}. SqPnP dramatically outperforms EPnP which performs between two and four times worse. Interestingly, we observe that EPnP shows a large but nearly constant orientation uncertainty after \unit[2]{m}, whereas SqPnP shows a linear increase in the noise standard deviation. Note that we do not compare against EPnP with nonlinear refinement as the OpenCV implementation requires at least six points for iterative refinement.

\subsection{Closed-Loop Deployment}

\begin{figure}[t]
    \centering
    \input{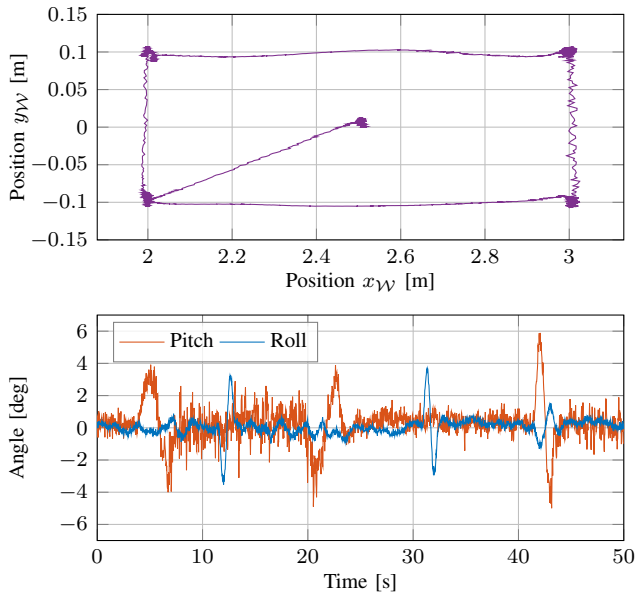}
    \vspace*{-18pt}
    \caption{Closed-loop experiments: the drone flies a rectangular pattern starting at (2,0.1) and then lands at a distance of \unit[2.5]{m}. The event-camera is located at the origin of the coordinate system at $(x_\wfr, y_\wfr) = (0,0)$ and the optical axis of the \unit[50]{mm} lens is aligned with the $x_\wfr$ direction. The bottom plot shows the roll and pitch angle measurements during the flight.}
    \label{fig:closed_loop}
\end{figure}

In this experiment we fly the small drone shown in Fig.~\ref{fig:overview} in closed-loop and the event-camera motion capture runs at \unit[400]{Hz} leading to a system latency of \unit[2.5]{ms}. The monocular event-camera motion capture system is the only source of state-estimation for the MPC controller~\cite{foehn2022agilicious} of the quadrotor. The drone is tasked to fly a rectangular pattern, hover and then land. The trajectory is shown in the top plot Fig.~\ref{fig:closed_loop} and the roll and pitch angle measurements are shown at the bottom. When the drone is further away at {$x_\wfr=\unit[3]{m}$} the position and orientation estimate become more noisy but overall we find that the system is able to safely fly the small drone, thereby demonstrating the performance of the developed system.

\section{Implementation Details}
\label{sec:implementation_details}
To ensure real-time operation of the event-camera motion capture system the delay in the processing pipeline must be kept to a minimum. This is achieved through an efficient, multi-threaded implementation in combination with filtering data early on in the pipeline. This section gives an overview of the most important concepts, specifically the data representation, the filtering and the multi-threading.

\subsection{Event Data Representation}
\label{subsec:event_data_representation}
A single event as supplied by the camera is given as a four-tuple, consisting of an $x$ and a $y$ coordinate (both \texttt{uint16\_t}), a polarity $p$ which is either -1 or 1 (\texttt{int8\_t}), and a timestamp $t$ (\texttt{uint64\_t}). Additionally, 3 bytes of padding are included for 16-byte alignment. 
The event camera supplies a stream of such raw events. For the following discussion of different data representations for blinking LED detection, an event stream containing $k$ events over a time period $T$ coming from an event camera with image width $W$, height $H$ and $N = W \times H$ pixels is considered.

\subsubsection*{1D Representations}
The \emph{event stream} is the most basic and raw representation of events and has recently gained some attention in combination with spiking neural networks~\cite{gehrig2020eventbasedangular}. For LED detection with classical CPU architectures however this representation is completely unsuitable. To extract any spatial information, the entire event stream must be searched for pixels with matching coordinates. Furthermore, having a memory layout where each event is stored serially is not efficiently using the cache: if we search for a given x coordinate, the remaining 14 bytes of the raw event representation are unused and just occupy cache space.

\subsubsection*{2D Representations}
In the \emph{event frame} representation the events are stored as a 2D grid by either summing the polarity or by counting the number of events. The accumulation is done for the time window of length $t$ which represents the equivalent of the exposure time. The conversion from an event stream to an event frame is fast as can be done in linear time $\mathcal{O}(k)$ by iterating once over the stream. This representation is suitable for filtering out which pixels have a sufficiently high number of events to be candidates for a blinking LED, however it does not include any time information which would allow robust frequency detection.

A \emph{time-surface} representation is also a 2D image, but each pixel in this 2D grid is assigned the value of the latest timestamp. For detecting blinking LEDs this representation is unsuitable as it contains no information related to periodic on-off transitions of a pixels.

\subsubsection*{3D Representations}
In an \emph{event volume}, events are stored as a 3D grid in a form that can be thought of as a stack of multiple event frames. This representation also includes time information and, given a sufficiently fine binning in the time domain, could be used to detect the frequency of a blinking LED. However, to accurately detect the frequency the binning would have to be very fine, yielding a huge memory footprint. For an accuracy of \unit[5]{$\mu s$}, a window length $t = \unit[2]{ms}$, VGA resolution and \texttt{uint8\_t} storage, the event volume would occupy \unit[117]{MB} of memory. This size exceeds all levels of the processor cache, potentially affecting runtime adversely.

\subsubsection*{Signed Delta-Time Volume}
\begin{figure}
    \centering
    \input{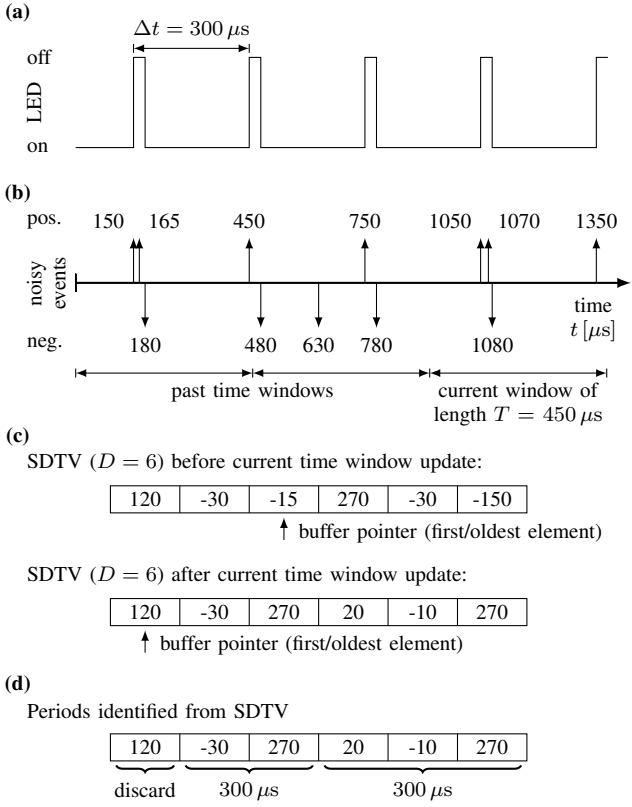}
    \vspace*{-18pt}
    \caption{Illustration on the construction of the \emph{Signed Delta-Time Volume (SDTV)} from an event stream. \textbf{a)} The LED is blinking with a period of $\unit[300]{\mu s}$ with a duty cycle of \unit[10]{\%}. \textbf{b)} A single pixel of the event camera records a noisy signal of this blinking LED. False double events (e.g. at $t = \unit[150]{\mu s}, \unit[165]{\mu s}$) and spurious events (e.g. at $t = \unit[630]{\mu s}$) are included. \textbf{c)} Construction of the SDTV illustrated before processing the latest time window and after processing the time window. \textbf{d)} Periods robustly identified from the SDTV by summing up absolute time differences between negative $\rightarrow$ positive transitions (the first positive value is included). All events until the first positive $\rightarrow$ negative transition are discarded.}
    \label{fig:sdtv}
\end{figure}

To get past the shortcomings of those widely used event representation we propose a data representation that is ideally suited for the task of blinking LED detection: the \emph{signed delta-time volume (SDTV)}. It is a 3D volume of size {$W \times H \times D$} where $D$ is the stack depth. For each pixel the time difference to the last event is stored and the polarity of the event is encoded in the sign of this time difference. This is possible because time must be monotonically increasing, so we can re-purpose the sign-bit for polarity encoding. The idea is illustrated in Fig.~\ref{fig:sdtv} for a single pixel stack of the SDTV.

Because of the fast blinking of the LEDs the time differences (in microseconds) between consecutive events are always within \texttt{int16\_t} range, making storage compact. As most operations are done per pixel-stack, the memory layout is such that the $D$-dimension is consecutive. Similar to the other representations, converting an event stream to SDTV is linear in the number of events.
The signed delta-time volume is not computed per window of length $T$ but updated as a cyclic buffer. This increases the accuracy of the frequency detection for LEDs blinking at a lower frequency than $f_\text{max}$ since the amount of LED periods available for frequency identification is independent of the frequency.

The minimal depth $D$ can be calculated based on the window length $T$ and the frequency of the fastest LED $f_\text{max}$ as $D = 2\,T\,f_\text{max}$ because every LED should trigger two events (once on and once off) per period. Typical values of $D$ are between 4 and 16, reducing the memory footprint by a factor of 25 to 100 compared to a event volume.

\subsection{Filtering}
When computing the signed delta-time volume representation from the event stream for a time window $t$ of events, we also compute an event frame based on the event count of each pixel. Only pixels with more than {$\beta \cdot 2 t f_\text{min}$} events are considered further where $\beta$ is the probability that a transition triggers an event. We use {$\beta = 0.8$} to purposely underestimate the detection probability.

For all selected pixels the SDTV is used to calculate mean, median and standard deviation of the period. The period is defined as the time between two on-events with at least one off-event in between as illustrated in Fig.~\ref{fig:sdtv}d). In agreement with~\cite{censi2013activeled} we find that this is a robust measure. After rejecting pixels with a too-large standard deviation in the period, pixels are clustered together. Too small and too large clusters are rejected as the expected size of an LED is roughly known a priori. Clusters are then assigned to the individual LEDs by matching the measured average period in a cluster with the blinking frequencies of the LEDs. Each LED is tracked by a particle filter that gets the assigned clusters for each LED as an input.

\subsection{Multi-Theading}
\label{subsec:multi_threading}
In a real-time application like this, relying on generic multi-threading tools such as OpenMP can be problematic. For this reason, the threading is manually implemented to ensure optimal performance. Each thread in the pipeline shares its memory with the next thread in the pipeline. To ensure threads do not block each other, each thread allocates the required memory two times. During operation, the thread writes to one of its allocations while the other memory chunk is processed by the next pipeline step. Subsequently, the memory pointers are swapped and the newly filled batch processed. 

The pipeline primarily consists of three threads. They
\begin{enumerate}
    \item copy events from event camera driver into a buffer,
    \item convert a linear event buffer into the optimized SDTV representation described in Section~\ref{subsec:event_data_representation}, and
    \item process the accumulated data to detect the LEDs, assign the LED clusters and solve the PnP (perspective-n-points) problem.
\end{enumerate}

This design makes it possible to run the pipeline at hight speeds on a modern laptop. Speeds exceeding \unit[1]{kHz} are possible, but due the slowest LED blinking at \unit[1700]{Hz} increasing the processing frequency beyond \unit[800]{Hz} might degrade robustness.
\section{Blinking LED Circuit}
\label{sec:blinked_led_circuit}
\begin{figure*}
    \centering
    \includegraphics{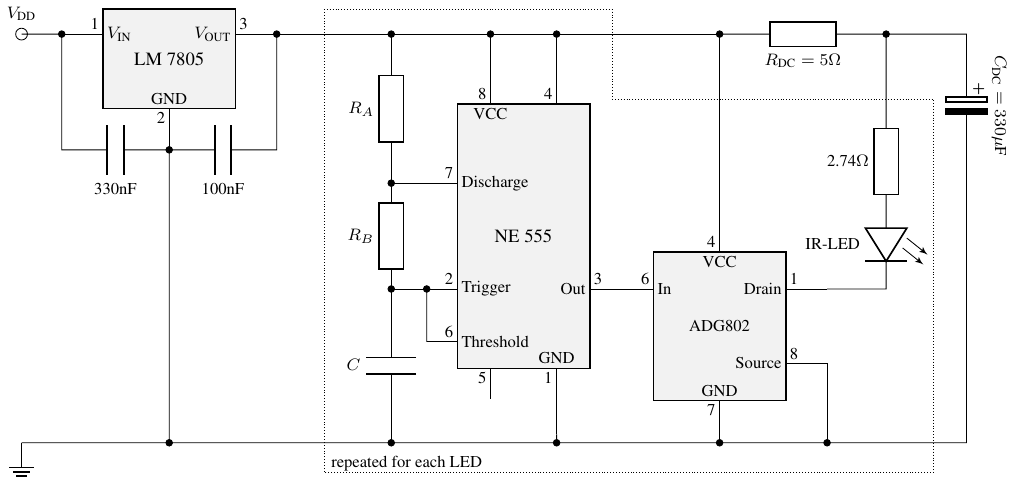}
    \caption{Circuit diagram of the complete blinking LED circuit. For simplicity we only show one LED driver circuit (area enclosed in dotted line), which is then replicated multiple times to drive multiple LEDs. The input voltage range for the power supply is $V_\text{DD}$ is \unit[8-35]{V}. The resistor and capacitor values $R_A$, $R_B$, and $C$ for the A-stable NE555 operation at different frequencies are given in Tab.~\ref{tab:resistor_capacitor_values}. For improved clarity the standard \unit[100]{nF} ceramic bypass capacitors to stabilize VCC for the NE555 and ADG802 have been omitted in the circuit diagram above.}
    \label{fig:circuit_diagram}
\end{figure*}
\subsection{Choice of LEDs}
\label{subsec:choice_of_leds}
For the accuracy of the event-based mocap system it is important that the center of the blinking LED can be detected easily by the event camera. To achieve this, the LED should be small, very bright and have a short switching time to produce a well-defined rising and falling edge. These requirements are ideally met by LEDs optimized for pulsed operation, such as infrared LEDs for data transmission.

In this work, the Osram SFH4350~\cite{osram4350} IR-LED \unit[3]{mm} was selected as it features an extremely short switching time of \unit[12]{ns}. Being designed for opto-electronics it also permits up to \unit[1]{A} pulsed forward-current for pulses shorter than \unit[100]{$\mu$s} if the duty cycle is below \unit[2]{\%}. The LED has its peak emission around \unit[850]{nm} with a spread of \unit[50]{nm} which is within the sensitivity of most CMOS-based imaging sensors.

\subsection{Circuit Design}
\label{subsec:circuit_design}

The design of the LED driver circuit is tightly coupled with the entire event-based motion capture system:
\begin{enumerate}
    \item Faster blinking frequencies increase the reactiveness of the system as at least one full period must be detected for identifying an LED. For robustness, a more conservative approach to detect at least two periods is better. Consequently, if the LEDs blink at \unit[1]{kHz} the overall system is limited to \unit[500]{Hz} output rate.
    \label{item:lower_limit}
    \item If the LEDs blink too fast, measuring the signal becomes difficult. Typically event cameras perform very well at measuring signals with frequencies up to \unit[2-3]{kHz}~\cite{wang2024towardshighspeed, crabtree2023refactoryperiod, lichtsteiner2008a128x128dvs} and the events are timestamped with \unit[1]{$\mu$s} time resolution.
    \label{item:upper_limit}
    \item At least four LEDs are neccessary to yield a unique solution to the PnP problem.
    \label{item:number_leds}
    \item the individual frequencies should not alias into each other. This means that, ideally, all LEDs have blinking frequencies within a factor of two.
    \label{item:aliasing}
    \item Due to the limitations of the LED a duty cycle of \unit[2]{\%} can not be exceeded.
    \label{item:duty}
\end{enumerate}

To control the blinking LED either a microcontroller or an analog circuit can be used. Because the high LED forward current of \unit[1]{A} necessitates an analog output stage, we opted for a fully analog design using NE555~\cite{ne555} precision timers. To generate the signal for the LEDs, the NE555 is operated in A-stable mode (c.f. Sec. 8.3.2~\cite{ne555}). 

The current output of the precision timer is limited to \unit[200]{mA}, but its performance significantly degrades if the output current exceeds \unit[10]{\%} of the maximum value (c.f. Figure 3 of~\cite{ne555}). Therefore an SPST (single pole, single throw) digital switch is used. We selected an ADG802 as it features close to \unit[1]{A} pulsed current and has typical switching times around \unit[55]{ns}~\cite{adg802}. While considerably slower than the LED, it is still fast enough for the given application.

\begin{table}[t]
    \centering
    \caption{\vspace*{6pt}\parbox{1\linewidth}{\textnormal{Resistor and capacitor values (see Fig.~\ref{fig:circuit_diagram}) for the different LEDs. The calculated periods as well as the measured frequencies $f$ and duty-cycles $\alpha$ are listed.}}}
    \vspace*{-6pt}
    \label{tab:resistor_capacitor_values}
    \begin{tabularx}{1\linewidth}{CCC|CCC|CC}
\toprule
\multicolumn{3}{c|}{Part Specification} & \multicolumn{3}{c|}{Calc. from Sec. 8.3.2 \cite{ne555} } & \multicolumn{2}{c}{Measured} \\[2pt]
$R_A$ \newline [$\unit{k\Omega}$] 
& $R_B$ \newline [$\unit{k\Omega}$] 
& $C$ \newline [\unit{nF}] 
& $t_\text{on}$ \newline [$\unit{\mu s}$] 
& $t_\text{off}$ \newline [$\unit{\mu s}$] 
& $f$ \newline [\unit{kHz}]
& $f_\text{meas}$ \newline [\unit{kHz}] 
& $\alpha_\text{meas}$ \newline [\%]\\ \midrule %
68.1 & 0.39 & 10 & 2.7 & 477 & 2.094 & 1.73 & 0.66 \\
59.0 & 0.39 & 10 & 2.7 & 415 & 2.413 & 1.98 & 0.75 \\
51.1 & 0.39 & 10 & 2.7 & 359 & 2.781 & 2.29 & 0.87 \\
44.2 & 0.39 & 10 & 2.7 & 312 & 3.207 & 2.61 & 0.99 \\
40.2 & 0.39 & 10 & 2.7 & 284 & 3.520 & 2.86 & 1.09 \\
\bottomrule
\end{tabularx}
\end{table}

The LEDs, precision timers and switches are all supplied with a single LM7805 voltage regulator. This is possible because the time-averaged load is well below the design limit of the voltage regulator. Each NE555 draws {$I_\text{NE555} = \unit[3]{mA}$} of supply current. The time averaged current $\bar{I}_\text{LED}$ for an LED pulsed with a duty cycle $\alpha = t_\text{on} / t_\text{period}$ with a pulse current {$I_p = \unit[1]{A}$} is given by
\begin{equation}
    \bar{I}_\text{LED} = \alpha \cdot I_p
\end{equation}
Assuming there is $N=5$ LEDs and they are operated at an average duty cycle of \unit[1]{\%} this leads to a total, time averaged current of 
\begin{equation}
    \bar{I} = N \cdot \left(\bar{I}_\text{LED} + I_\text{NE555} \right) = \unit[65]{mA}
\end{equation}
which is within specifications for an LM7805~\cite{lm7805} without any additional cooling (given the TO220 package and a \unit[12]{V} supply). Therefore, a sufficiently large decoupling capacitor $C_{DC}$ charged through $R_{DC}$ is used to supply the LEDs with power, effectively shielding the LM7805 from all current spikes caused by the LEDs. Based on all the above considerations, the circuit shown in Fig.~\ref{fig:circuit_diagram} has been designed and subsequently manufactured into a PCB with SMD version of the NE555 and the ADG802. The dotted line marks the components in the circuit that are replicated for each LED.

\begin{figure}
    \centering
    \input{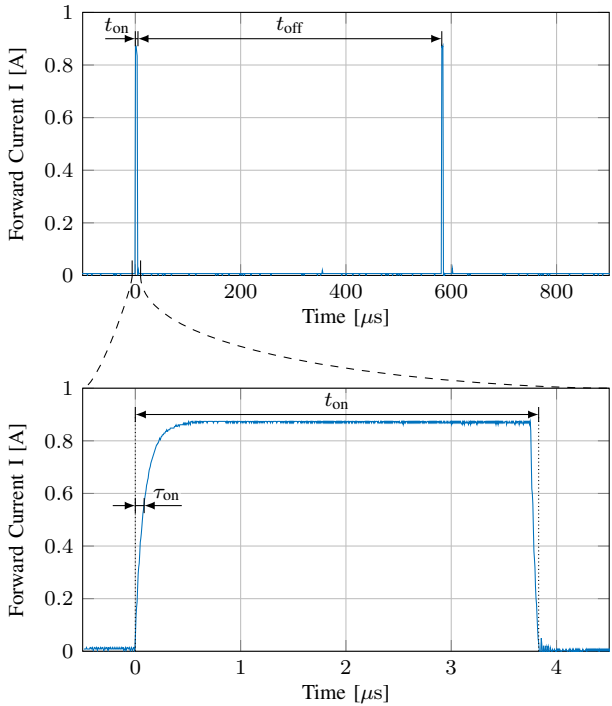}
    \vspace*{-6pt}
    \caption{Current through an LED blinking at the lowest frequency of \unit[1.73]{kHz}. The upper plot shows the entire period $t = t_\text{on} + t_\text{off}$, whereas the lower plot shows only the pulse through where the LED is on for $\unit[3.8]{\mu s}$. From the plot, we get a switch-on time constant $\tau_\text{on}$ of the LED of \unit[84]{ns} (time to reach \unit[63]{\%} of the steady-state value).}
    \label{fig:led_current}
\end{figure}

The resistor and capacitor values used in the NE555 timer circuit are listed in Tab.~\ref{tab:resistor_capacitor_values}. The values have been calculated such that the frequencies of all 5 LEDs follow the points \ref{item:upper_limit} to \ref{item:duty}. After building and manufacturing of the PCB, the measured frequencies are also listed in Tab.~\ref{tab:resistor_capacitor_values}. Note that the mismatch w.r.t the calculated values is about \unit[20]{\%}. This mismatch is consistent across 5 identical copies of the board and of no concern for the practical applications as it is straightforward to measure the blinking frequency with an oscilloscope. Exemplarily, Fig.~\ref{fig:led_current} shows the current through one LED.

\section{Conclusion}
\label{sec:conclusion}
In this technical report we present that a low-latency, high-accuracy monocular motion-capture system with an event-camera. The experiments with static objects show that the system has millimeter accuracy and the closed-loop experiments with the small quadrotor demonstrate that it is is well-suited for real-time control in mobile robotics tasks. 

Originally, we developed the system with a focus on confined environments where a commercial, multi-camera system can not be used. However, we now believe that monocular event-camera motion-capture systems are highly relevant in a broader sense: event-cameras are expensive sensors, but the cost of a single event camera is 10 to 100 times less than that of a full motion-capture system, making the presented approach appealing for low-cost applications where the limited tracking volume (only in front of the event-camera) is a great trade-off for the small form factor, portability and cost.

{\small
\bibliographystyle{IEEEtran}
\bibliography{references}
\balance
}

\end{document}